\documentclass[11pt]{article}

\usepackage[final]{acl}
\usepackage[most]{tcolorbox}
\usepackage{times}
\usepackage{latexsym}
\usepackage{listings}
\usepackage{multirow}
\usepackage[T1]{fontenc}

\usepackage[utf8]{inputenc}

\usepackage{microtype}

\usepackage{inconsolata}

\usepackage{graphicx}

\title{Modeling Turn-Taking with Semantically Informed Gestures}

\author{
 \textbf{Varsha Suresh\textsuperscript{1}\thanks{These authors contributed equally to this work.}}, \textbf{M. Hamza Mughal\textsuperscript{2}}\footnotemark[1],
 \textbf{Christian Theobalt\textsuperscript{1,2}}, \textbf{Vera Demberg\textsuperscript{1,2}}
\\
\\
 \textsuperscript{1}Saarland University,
 \textsuperscript{2}Max Planck Institute for Informatics, Saarland Informatics Campus
\\
\small{
\textbf{Correspondence:}
 \texttt{\{vsuresh,vera\}@lst.uni-saarland.de}},
 \small{\texttt{\{mmughal,theobalt\}@mpi-inf.mpg.de}}
}

\begin{document}
\maketitle
\begin{abstract}
In conversation, humans use multimodal cues, such as speech, gestures, and gaze, to manage turn-taking. While linguistic and acoustic features are informative, gestures provide complementary cues for modeling these transitions. To study this, we introduce DnD Gesture++, an extension of the multi-party DnD Gesture corpus enriched with 2,663 semantic gesture annotations spanning iconic, metaphoric, deictic, and discourse types. Using this dataset, we model turn-taking prediction through a MoE framework integrating text, audio, and gestures. Experiments show that incorporating semantically guided gestures yields consistent performance gains over baselines, demonstrating their complementary role in multimodal turn-taking.

\end{abstract}

\section{Introduction}





Multi-party conversations are rich, dynamic interactions in which participants coordinate through both verbal and non-verbal channels. A fundamental mechanism that structures these interactions is turn-taking, the implicit management of who speaks next, when a speaker should continue, and when they should yield the floor \cite{sacks1974simplest}. Effective turn-taking relies on anticipating when a speaker will finish and when another should begin, a problem that becomes increasingly challenging as the number of participants grows.

Existing computational models of turn-taking have predominantly relied on verbal cues, such as lexical content~\cite{ekstedt2020turngpt} and prosody~\cite{ekstedt2022much}. However, human communication extends far beyond speech. Social interaction is inherently multimodal, with non-verbal behavior such as gesture and gaze, playing a crucial role in signaling intentions to hold or yield turns~\cite{duncan1979strategy,skantze2021turn}. Relying solely on text and audio therefore overlooks important communicative signals. Among non-verbal behaviors, semantic gestures (iconic, deictic, metaphoric, and discourse gestures) are especially informative because they are shaped by conversational meaning rather than prosodic rhythms~\cite{mcneill1992hand}. These gestures help structure dialogue, convey contextual information, and regulate the flow of interaction~\cite{bavelas1995gestures,holler2018processing}. For this reason, we focus on studying how semantic gestures contribute to turn-taking dynamics in multi-party settings and how incorporating them can improve predictive models.



\begin{figure}
    \centering
\includegraphics[width=\linewidth]{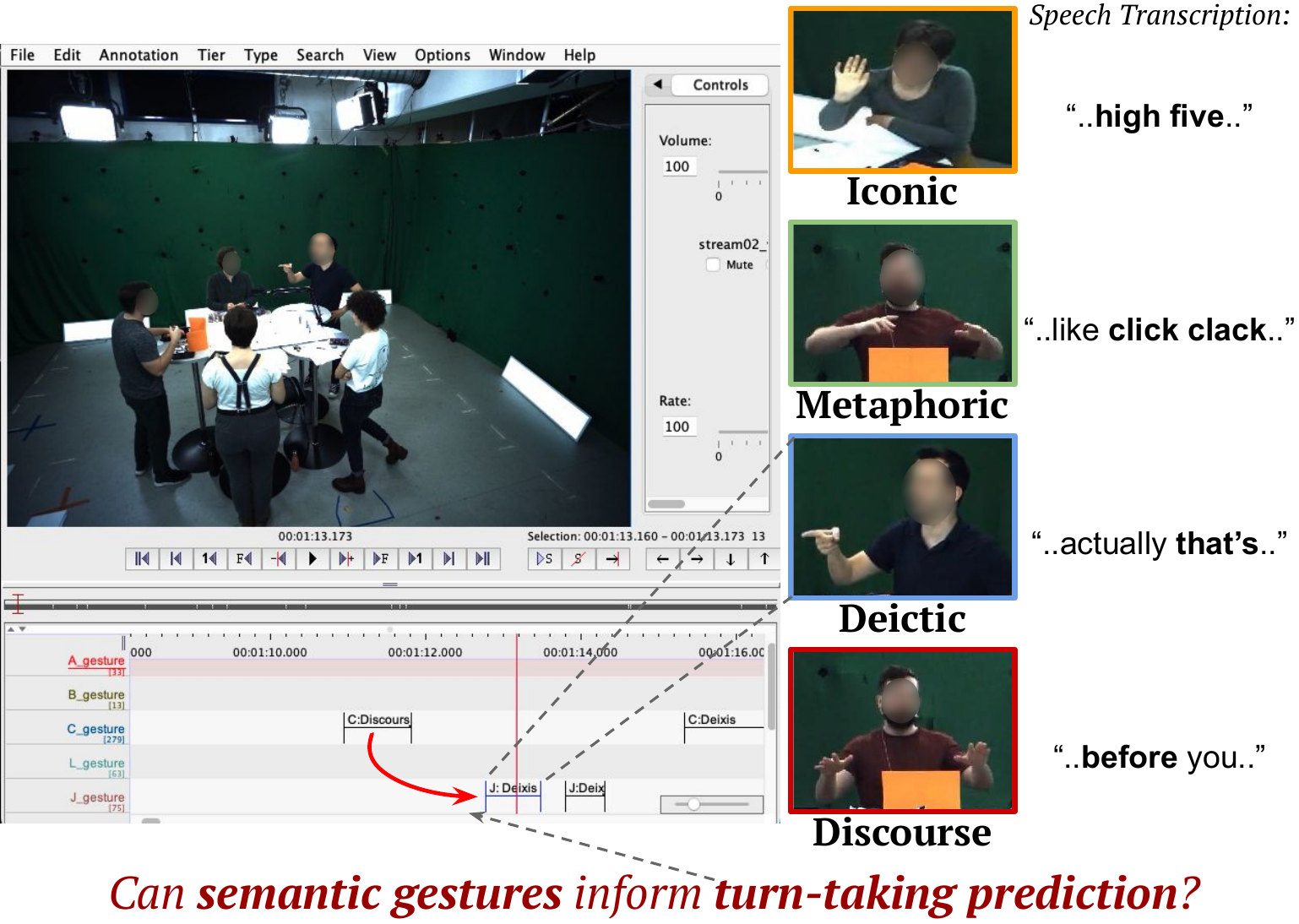}
    \caption{\textbf{Gesture Type Annotations.} Time-aligned labels contain semantic gesture types, that are determined by speech context. These labels can help learn gesture representations which improve turn-taking prediction in a multi-party conversation.}
    \vspace{-1em}
    \label{fig:elan}
\end{figure}
To investigate the role of semantic gestures in turn-taking prediction, we build upon the DnD Gesture dataset~\cite{mughal2024convofusion}, a multi-party conversational corpus containing synchronized 3D motion, audio, and transcripts from participants in a tabletop game. We manually annotated the six-hour corpus with gesture-type labels following \citet{mcneill1992hand}. The resulting DnD Gesture++ \footnote{Located within DnD Gesture dataset: \url{https://doi.org/10.17617/3.IPFYCC}} corpus includes 2,663 gesture instances across six hours (444 labels/hour), forming the most densely annotated English dataset of its kind. For comparison, BEAT~\cite{liu2024emage}, one of the few large-scale datasets with semantic gesture-type annotations contains approximately 288 labels per hour. The higher annotation density in DnD Gesture++ provides richer coverage of gesture behavior and more fine-grained supervision for downstream multimodal modeling tasks. We further reformat this data for turn-taking prediction task, labeling 12k turns as either \textit{hold} or \textit{yield}. 

In this work, we model turn-taking using a Mixture-of-Experts (MoE) framework that fuses text, audio, and gesture modalities through a gating network. Gesture embeddings are semantically enriched using our annotations to better align motion cues with linguistic and acoustic information. Our experiments demonstrate consistent gains from incorporating gestures, particularly when gesture embeddings are semantically supervised. We further analyze the latent space and modality weights from the MoE framework to better understand the contribution of semantic gesture representations. Beyond turn-taking, the dense annotations in DnD Gesture++ provide a valuable resource for tasks such as co-speech gesture generation, enabling more semantically grounded gesture synthesis~\cite{kucherenko2021speech2properties2gestures,mughal2025raggesture}.

\section{Related Work}

\subsection{Modeling Turn-Taking for Spoken Dialog }

The coordination of speaking turns in human dialogue is inherently multimodal \cite{skantze2021turn}. Speakers use non-verbal signals such as prosody, gaze, and hand gestures to project turn boundaries and manage conversational flow \cite{duncan1979strategy,kendrick2023turn}. Gestures serve key semantic functions: pragmatic or discourse-related gestures often signal a \textit{yield}, while representational gestures like iconic forms indicate to \textit{hold} the floor \cite{bavelas1995gestures,holler2018processing}. Moreover, they serve as reliable cues for turn-taking and help maintain conversational smoothness \cite{holler2018processing,ter2024gestures,kendrick2023turn,holler2019multimodal,hofstetter2021achieving}.

Early dialogue systems detected turn boundaries using fixed silence thresholds, while recent data-driven approaches use syntactic and pragmatic cues from dialogue transcripts~\cite{ekstedt2020turngpt}. Multimodal approaches further integrate prosody \cite{ekstedt2022much}, face features \cite{russell2025visual,lin2025predicting,kurata2023multimodal} to enhance accuracy. However, the effect of semantic gestures remains underexplored in turn-taking models. Our work examines whether semantic gesture types enhance turn-taking prediction in language models.

\subsection{Role of Semantic Gestures in Spoken Dialog}

Co-speech gestures are typically classified as rhythmic (beat) gestures, aligned with prosody, or semantic gestures, aligned with meaning \cite{mcneill1992hand,nyatsanga2023comprehensive}. Semantic gestures play a key role in communication by enhancing comprehension \cite{holler2018processing} and clarifying speaker intent \cite{goldin2014gesture}. They also provide valuable signals for tasks like multimodal reference resolution \cite{ghaleb2025see}, discourse marker disambiguation \cite{suresh2025enhancing}, and co-speech gesture synthesis~\cite{mughal2025raggesture,kucherenko2021speech2properties2gestures, ram2025gesturecoach}.

Semantic gestures are often classified by McNeill’s taxonomy \cite{mcneill1992hand}, which reflects their communicative intent. The main types include iconic, metaphoric and deictic gestures. Discourse gestures that structure dialogue, such as signaling topic shifts, are also considered semantic \cite{bavelas1995gestures}. Refer to Sec.~\ref{subsec:annotation} for details.
Datasets like SAGA \cite{lucking2013data,kucherenko2021speech2properties2gestures} and BEAT \cite{liu2022beat} include semantic type labels but are limited to scripted or monadic data \cite{lucking2013data,liu2022beat}. The DnD Group Gesture Dataset \cite{mughal2024convofusion} captures natural multiparty interaction but lacks such labels. Our work extends it with semantic type annotations, yielding the first multiparty dataset for studying gesture-informed turn-taking modeling.

\section{DnD Group Gesture++}

The DnD Group Gesture Dataset~\cite{mughal2024convofusion}
captures co-speech gestures in multi-party conversations during a Dungeons \& Dragons (DnD) roleplaying game. It includes full-body motion and fine-grained finger articulation from five English-speaking participants over four sessions (6 hours total).

\subsection{Gesture type annotation}
\label{subsec:annotation}
We extend the DnD Gesture \cite{mughal2024convofusion} with gesture type annotations based on McNeill’s framework~\cite{mcneill1992hand, mcneill2005gesturethought}, classifying gestures as iconic, metaphoric, deictic, or discourse~\cite{goldin1993transitions, kendon1995gestures, khosrobeigi2022gesture}.
\begin{table}[]
\resizebox{\linewidth}{!}{
\begin{tabular}{ccccc}
\hline \hline
\textbf{Overall} & \textbf{Deixis} & \textbf{Metaphoric} & \textbf{Iconic} & \textbf{Discourse} \\ \hline
0.522  & 0.576 & 0.458  & 0.674  & 0.081  \\  \hline \hline 
\end{tabular}}

\caption{Interrater agreement scores}
\label{tab:kappa}
\end{table}
The annotations include (\# of samples), \textbf{iconic} (724) gestures, which depict concrete objects; \textbf{metaphoric} (151) gestures, which represent mental images of abstract concepts; \textbf{deictic} (1155) gestures, which involve referential gestures like pointing; and \textbf{discourse} (633) gestures, which are elicited by the structure of the spoken discourse. We focus on these semantic gesture types as they capture meaning-related aspects of communication that extend beyond prosodic or lexical information and may therefore provide complementary signals for turn-taking. Beat gestures are not included, as their close association with prosody overlaps with cues already available in the audio modality. Table~\ref{tab:kappa} reports interrater agreement, with an average Cohen’s $\kappa$ of 0.52. It also presents per-class $\kappa$ scores, computed by treating each gesture category as a binary classification task, where the target class is considered positive and all others negative. Agreement is notably lower for discourse gestures, which are more challenging to annotate, as they are defined primarily by communicative function rather than by gesture form. Additional details on the annotation process are provided in Appendix~\ref{appx:annotation_details}.

\begin{table}[]
\resizebox{\linewidth}{!}{
\begin{tabular}{l|cccc|c}
\hline
\multirow{2}{*}{\textbf{Label}} & \multicolumn{4}{|c|}{\textbf{\% of gesture type}} & \multirow{2}{*}{\textbf{\begin{tabular}[c]{@{}c@{}} turns w sem gesture \\ (total \# of turns)\end{tabular}}} \\ \cline{2-5}
& \textbf{De}   & \textbf{Di}  & \textbf{Ic}  & \textbf{Me}  &                           \\ \hline
\textit{hold}                            & 42.3          & 23.1         & 28.6         & 5.9          & 1550 (7722)                \\
\textit{yield}                           & 44.4          & 24.2         & 25.7         & 5.7          & 990 (5087)               \\ \hline
\end{tabular}}
\caption{Distribution of semantic gesture types across turn-taking labels. For each turn type (hold or yield), the table shows the relative frequency (\%) of each gesture type, deictic (de), discourse (di), iconic (ic), and metaphoric (me), and the number of turns containing at least one semantic gesture, along with the total number of turns in that label.}
\label{tab:data_dist}
\end{table}
\subsection{Converting to turn-taking data}

To model turn-taking, we convert continuous multi-party conversations in DnD Gesture++ into a structured dataset of discrete prediction instances, following previous works \cite{kelterer2025turn,jokinen2013gaze,ekstedt2023automatic}. We segment audio into Inter-Pausal Units (IPUs), continuous speech segments bounded by silence, with each IPU ending at a Transition Relevance Place (TRP) where a speaker change may occur. Turns are labeled \textit{yield} if another participant speaks next, or \textit{hold} if the same speaker continues, using a 200 ms silence threshold~\cite{lee2023multimodal, hara2019turn}. Short IPUs are merged with adjacent utterances. This results in 12k turns (7k hold, 5k yield). We split the data into 70\% train, 10\% validation, and 20\% test. Refer to Table~\ref{tab:data_dist} for more details.

\section{Multimodal Modeling of Turn-Taking Prediction}

\begin{figure}[h]
    \centering
\includegraphics[width=\linewidth]{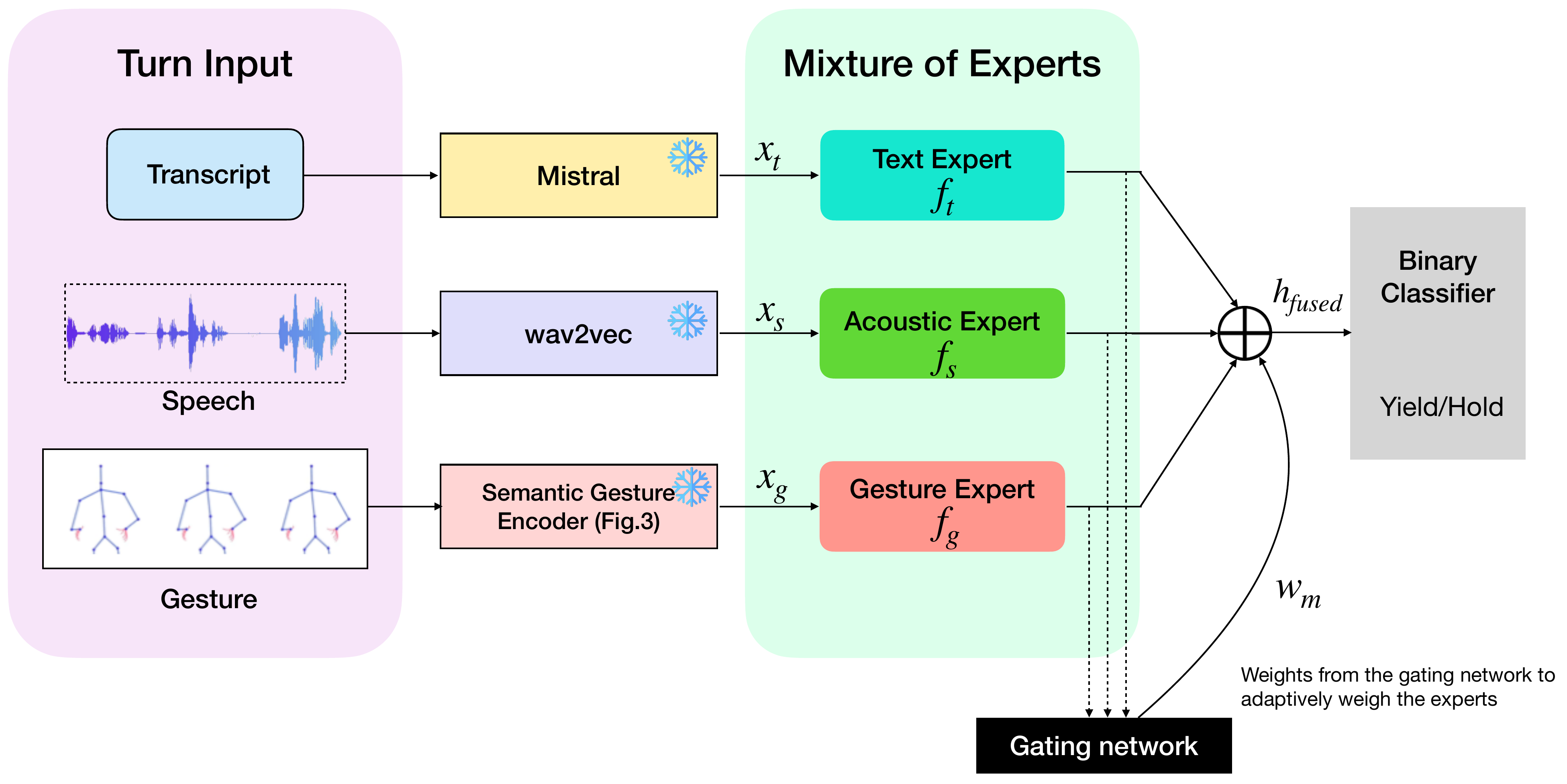}
    \caption{MoE modeling of turn taking}
    \vspace{-1em}
    \label{fig:model}
\end{figure}

We model turn-taking using text, audio, and gestures captured via motion data. Each modality is processed by a dedicated expert within a MoE framework, where the outputs of all experts are fused via a gating network that dynamically weighs each modality based on context. Let $x_{m}$ denote the input features for modality $m$ and $f_{m}(x_{m})$ its expert output. The gating network computes modality weights $w_{m}$ using softmax over concatenated weights. The fused representation is a weighted sum of expert outputs $h_{\text{fused}}=\sum_{m=1}^{M} w_{m} \cdot f_{m}(x_{m})$.

\begin{figure}[h]
\centering
\includegraphics[width=\linewidth]{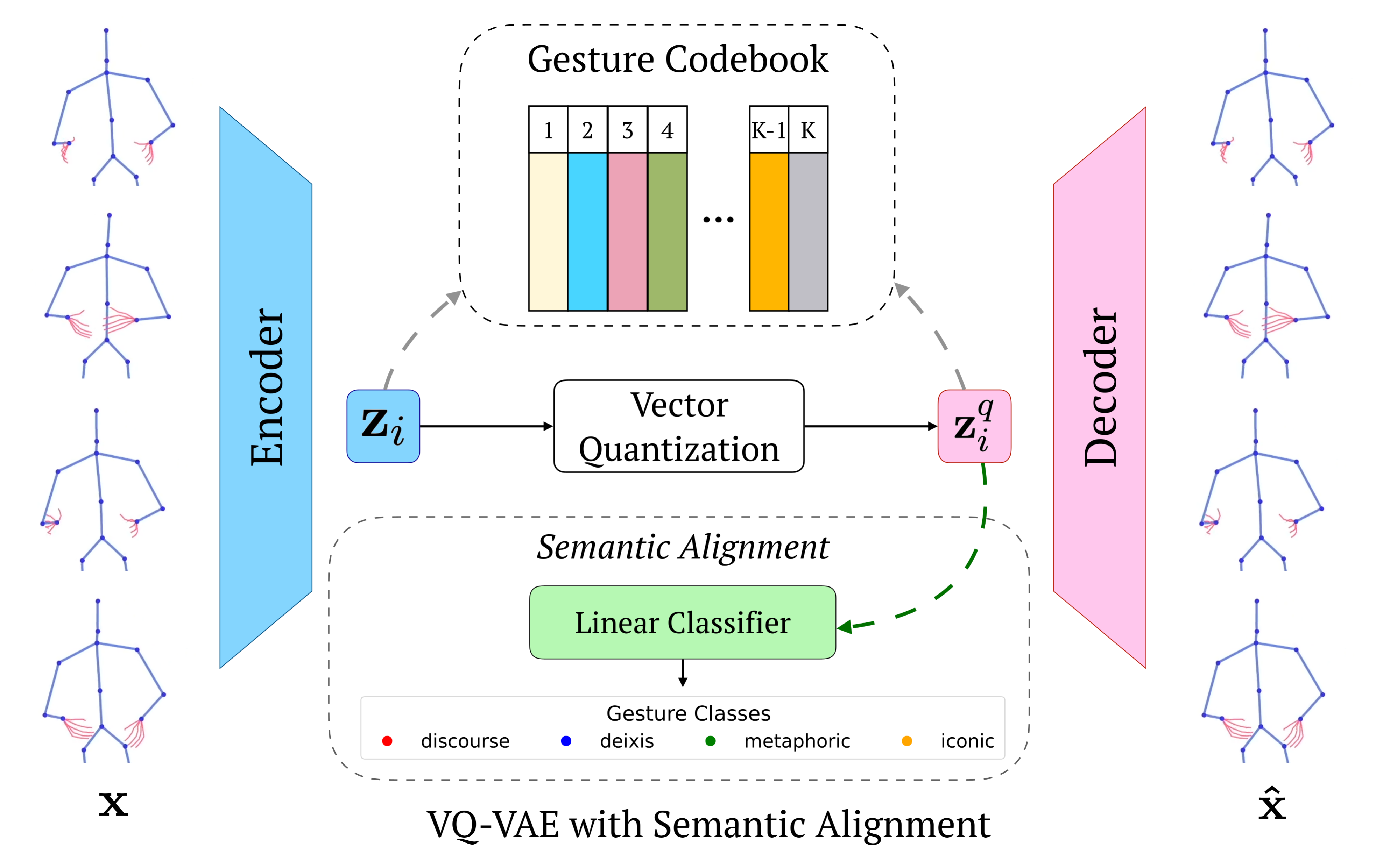}
    \caption{Learning semantically-aligned gesture representations.}
    \vspace{-1em}
    \label{fig:sem-vq}
\end{figure}

Finally, $h_{fused}$ is passed through a linear classifier to predict \textit{hold} or \textit{yield}. Text and audio features are obtained from pretrained embeddings, sentence embeddings for text\footnote{\url{https://huggingface.co/mixedbread-ai/mxbai-embed-large-v1}}
and Wav2Vec2 for audio\footnote{\url{https://huggingface.co/facebook/wav2vec2-large-960h}} and projected via MLP layers. 
Gestures are encoded following recent VQ-VAE based gesture tokenization approaches~\cite{liu2024emage, suresh2025enhancing}. Given a sequence of 3D upper body motion $\mathbf{x}$, the encoder network compresses it into a sequence of tokens. Each token $\mathbf{z}_i$ is then quantized via a codebook~\cite{van2017neural} to produce $\mathbf{z}^q_i$. The sequence of quantized embeddings is then used to reconstruct the input motion $\mathbf{x}$. Consequently, the VQ-VAE is trained through an MSE loss on reconstructed motion $\mathbf{\hat{x}}$ and input motion $\mathbf{x}$. Since the base model lacks semantics, we use gesture type annotations to inject semantic information (see Figure~\ref{fig:sem-vq}). For each turn, we obtain the gestures token sequences of codebook embeddings from the trained VQ-VAE which are then processed by a transformer encoder \cite{vaswani2017attention} which forms the gesture expert. The output of the Transformer is mean pooled over the sequence length to produce a fixed-length representation.


\section{Experiments}

Our work addresses two key research questions: (i) Does the inclusion of body movement features such as gestures improve turn-taking prediction beyond text and audio? (ii) Do semantically informed gesture representations offer advantages over raw motion features? 
 
\subsection{Model Comparisons}

We systematically compare our Text+Audio+Gesture model against single- and dual-modality baselines. We then evaluate the impact of semantic gesture embeddings from a VQ-VAE versus raw motion features. We also benchmark our MoE-based fusion against existing multimodal turn-taking approaches using the same modality experts for fair comparison. Importantly, our main aim is to assess the effect of gesture representations on turn-taking beyond text and audio, and the MoE setup allows us to quantify the contribution of each modality in this task.  Further implementation details can be found in the Appendix~\ref{appx:implementation}.

\begin{table}[h]
\centering
\resizebox{\linewidth}{!}{
\begin{tabular}{lcccc}
\hline \hline
 & \multicolumn{2}{c}{Overall} & \multicolumn{2}{c}{Per-class} \\ \hline
 & Acc & F1 & \textit{hold} & \textit{yield} \\ \hline
\textit{Majority Class} & 60.4 & 37.6 & 75.3 & 0.0 \\ \hline
Text & 68.8 $\pm {\scriptstyle{0.5}}$ & 66.6 $\pm {\scriptstyle{0.4}}$ & 75.1 & 58.2 \\
Audio & 67.1 $\pm {\scriptstyle{0.5}}$ & 63.2 $\pm {\scriptstyle{1.6}}$ & 75.1 & 51.3 \\
Gesture & 66.2 $\pm {\scriptstyle{0.2}}$ & 61.2 $\pm {\scriptstyle{1.6}}$ & 75.1 & 47.4 \\
Gesture (w/o sem) & 66.1 $\pm {\scriptstyle{0.3}}$ & 61.8 $\pm {\scriptstyle{0.5}}$ & 74.6 & 48.9 \\ \hline
Text+Audio & 69.7 $\pm {\scriptstyle{2.1}}$ & 67.9 $\pm {\scriptstyle{1.2}}$ & 74.9 & 61.1 \\
Text+Gesture & 70.0 $\pm {\scriptstyle{0.2}}$ & 67.8 $\pm {\scriptstyle{0.2}}$ & 76.2 & 59.4 \\
Audio+Gesture & 67.7 $\pm {\scriptstyle{0.4}}$ & 64.2 $\pm {\scriptstyle{0.2}}$ & 75.4 & 53.0 \\ \hline
Text+Audio+Gesture & \textbf{71.5 $\pm {\scriptstyle{0.3}}$} & \textbf{69.9 $\pm {\scriptstyle{0.1}}$} & \textbf{76.7} & \textbf{63.1} \\
\begin{tabular}[c]{@{}l@{}}Text+Audio+Gesture \\                      (w/o sem)\end{tabular} & 70.4$\pm {\scriptstyle{0.4}}$ & 68.7 $\pm {\scriptstyle{0.5}}$ & 75.9 & 61.5 \\ \hline \hline
\end{tabular}
}
\caption{Performance of multimodal variations for turn-taking prediction using text, audio, and gesture modalities with MoE based fusion modeling.}
\label{tab:main_res}
\end{table}

\begin{table}[]
\centering
\begin{tabular}{l|cc} \hline
Gesture Type & \textit{hold} & \textit{yield} \\ \hline
Discourse    & 78.3          & 74.1           \\
Deixis       & 78.2          & 65.4           \\
Iconic       & 72.5          & 58.3           \\
Metaphoric   & 89.5          & 66.7           \\
Overall      & 76.7          & 63.1          \\ \hline
\end{tabular}

\caption{Per-class F1 contributions for \textit{hold} and \textit{yield} predictions across each semantic gesture type}
\label{tab:per-class}
\end{table}

\begin{table}[h]
\centering
\resizebox{\linewidth}{!}{
\begin{tabular}{lccc}
\hline
& Overall       & \textit{hold}          & \textit{yield}       \\ \hline

ConcatFusion \cite{kurata2023multimodal} & 68.2          & \textbf{78.5} & 58.0          \\
LMF \cite{lin2025predicting}    & 67.9          & 77.2          & 58.7          \\
MoE (Ours)                                     & \textbf{69.9} & 76.7          & \textbf{63.1} \\ \hline
\end{tabular}
}			

\caption{Ablation study on fusion techniques: F1 scores of the Text+Audio+Gestures model evaluated with different existing multimodal fusion methods for turn-taking prediction.}
\label{tab:comparison_res}
\end{table}

\subsection{Turn-taking Prediction}

From Table~\ref{tab:main_res}, text performs best among single-modality experts, followed by audio. Gesture only models show lower performance, with little difference between non-semantic and semantic variants. This result reflects the nature of turn-taking cues—gestures complement rather than replace lexical and prosodic structure. Importantly, gestures consistently improve performance in fusion settings, indicating that they provide important information, especially for subtle transitions where speech cues are weak or ambiguous.
Multimodal fusion outperforms single-modality models with Text+Audio+Gesture model achieving the highest overall macro-F1 compared to the Text+Audio variant (p < 0.05). Notably, the semantically aligned gesture representation outperforms the non-semantic one (p < 0.05), showing that semantic supervision enhances multimodal fusion. 


When ablating with fusion techniques from prior multimodal turn-taking studies (Table~\ref{tab:comparison_res}), we observe per-class F1 imbalances, with the minority class particularly affected. In contrast, MoE-based fusion with semantically aligned gestures yields more balanced predictions across classes. Future work can explore fusion strategies to further improve alignment between text, audio, and gestures. Table \ref{tab:per-class} reports the per-class F1 scores for \textit{hold} and \textit{yield} predictions across semantic gesture categories. The results further demonstrate that different gesture types provide distinct cues for turn-taking: discourse gestures strongly contribute to \textit{yield} transitions, whereas metaphoric gestures are more indicative of \textit{hold} behavior. All results are averaged over three random seeds.

\subsection{ Analysing gesture representations}

\begin{figure}[h]
\centering
\includegraphics[width=\linewidth]{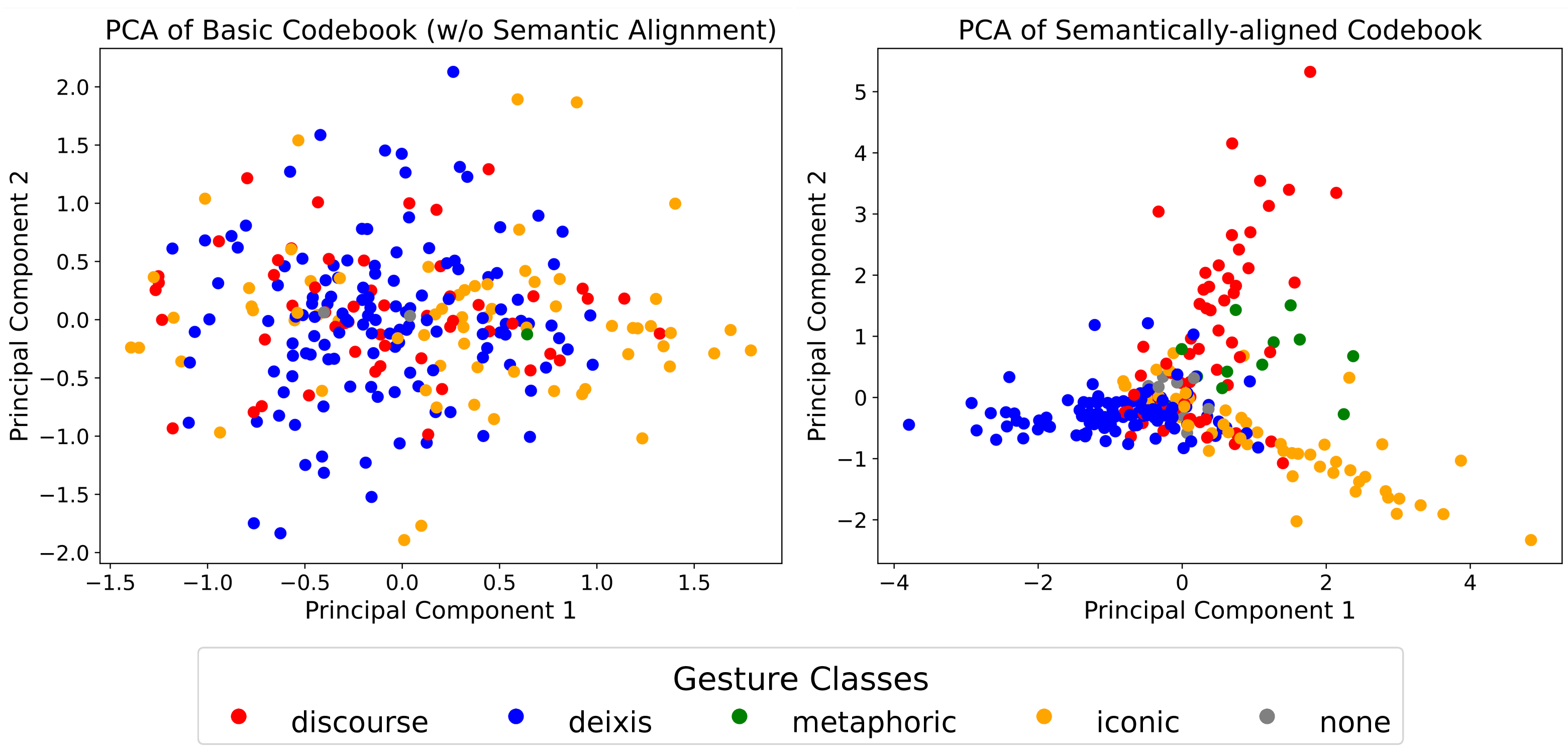}
    \caption{
    Visualization of Gesture Representations.}
    \vspace{-1em}
    \label{fig:sem-vq-pca}
\end{figure}

\begin{figure}[h]
    \centering
    \includegraphics[width=0.7\linewidth]{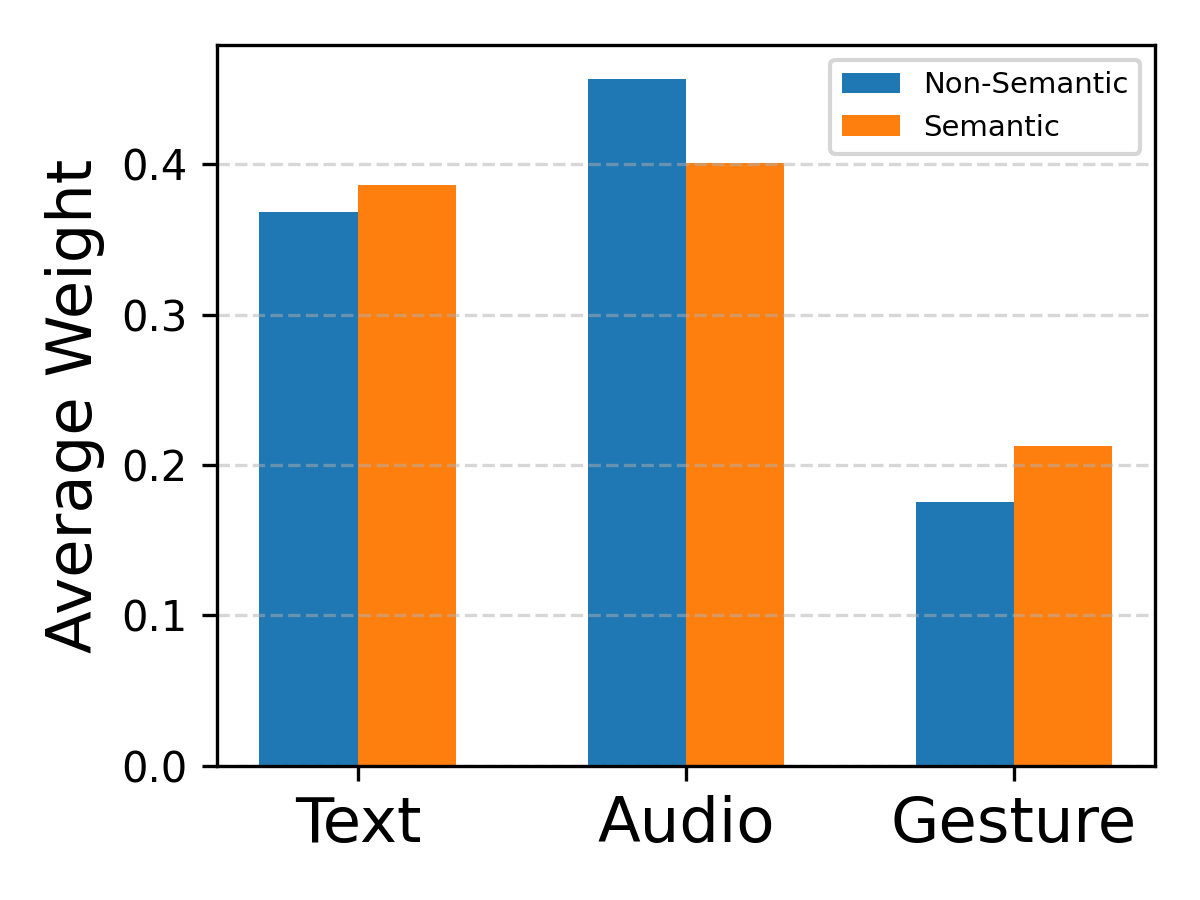}
    \caption{MoE modality weight contributions for semantic vs w/o semantic gesture representations}
    \vspace{-1em}
    \label{fig:moe_weights}
\end{figure}

Figure~\ref{fig:sem-vq-pca}, we visualize the gesture embeddings with and without semantic supervision. The semantically aligned embeddings form more distinct clusters corresponding to different gesture types, which facilitates better alignment with text and audio modalities. Figure~\ref{fig:moe_weights} further shows the average weight contribution for each modality in the MoE framework for the test samples. Using semantically aligned gestures increases the weights for both gestures and text, likely due to improved cross-modal alignment, which in turn supports better overall task performance.

\section{Conclusion} 
In this work, we investigate the role of semantic gestures in multimodal turn-taking prediction. We extended the multi-party DnD Gesture corpus with semantic annotations to create DnD Gesture++. Our results show that semantically guided gestures improve turn-taking modeling, especially when fused with text and audio. Beyond turn-taking, the corpus supports a range of multiparty interaction tasks, including discourse-level gesture analysis, co-speech gesture generation, and listener backchannel prediction, while addressing a key gap in semantic gesture resources for multi-party settings.

\section*{Limitations}

While this work is based on a naturalistic, game-based interaction setting, the domain specificity may limit generalisability. Future work should assess the role of semantic gestures in other interaction contexts, including task-oriented dialogues, meetings, remote communication, and cross-cultural settings. Although our semantic gesture annotations are carefully curated, they show only moderate inter-annotator agreement, reflecting the inherent subjectivity of gesture interpretation, particularly for discourse gestures. This annotation noise may cap achievable performance and could be mitigated through additional annotators. Our study focuses on whether semantic gestures provide complementary information for IPU-based turn-taking when combined with text and audio. While we employ a MoE fusion framework, future work could investigate more expressive fusion and temporal modeling approaches that jointly capture modality interactions and turn structure.




\section*{Acknowledgments}
This work was supported by the Deutsche Forschungsgemeinschaft, Funder Id: \url{http://dx.doi.org/10.13039/501100001659}, SFB 1102: ``Information Density and Linguistic Encoding'', project number 232722074, and funded in part by the Deutsche Forschungsgemeinschaft (DFG, German Research Foundation) -- GRK 2853/1 “Neuroexplicit Models of Language, Vision, and Action” - project number 471607914. We would like to thank Lucia Donatelli for her contribution to a previous version of the project. We extend our thanks to the annotators, Duc Anh and Anna Taylor, for their contributions.

\bibliography{custom2,custom}

\appendix

\section{Appendix}
\label{sec:appendix}

\subsection{Additional Annotation Details}
\label{appx:annotation_details}

The dataset contains 4 recording sessions. To label the gesture type in each recording, the annotators utilize the speech from the group conversation and multi-view videos to identify who made the gesture and mark it with label for the duration of said gesture. Therefore, the annotators considered both linguistic and visual modalities to identify the gesture types. Annotation was performed in ELAN \footnote{\url{https://archive.mpi.nl/tla/elan}} where each recording participant was assigned a separate track for labeling. For annotation, two English-speaking young students were recruited from the university. After that, authors randomly sampled the annotations and verified the annotations manually and cleaned conflicting or wrong annotations. To acquire the transcriptions we use, 
WhisperX \footnote{\url{https://github.com/m-bain/whisperX}}. 


\subsection{Implementation Details}
\label{appx:implementation}

For obtaining gesture representations: 
In order to train the VQ-VAE, we pass 3D upper body motion encoded as joint positions.
The input skeleton is normalized relative to pelvis (root) joint and translation is fixed to zero, such that the network only models hand and arm movements.
We utilize convolutional encoder and decoder networks based on ResNet backbone~\cite{he2016deep}.
To apply semantic alignment loss, we utilize the annotations by training using 4 annotated semantic classes i.e. iconic, deictic, metaphoric, discourse, and a "none" class consisting of beat and other non-semantic gestures.
During training, cross entropy loss is applied on the linear classifier to identify gesture categories.
This loss ignores the "none" class.
The whole framework is trained for 120 epochs with following hyperparameters: number of residual blocks = 2 in both encoder and decoder, embedding dim = 256, codebook size = 256, learning rate = 3e-4, optimizer = AdamW, reconstruction loss weight = 1, semantic loss weight = 0.1, and batch size = 512.

For gesture transformer encoder, we use a 1-layer transformer with a feedforward dimension of 128 and 4 attention heads. The final hidden states are mean-pooled to obtain the gesture expert representation. MoE training is conducted on a single NVIDIA V100 GPU with a batch size of 32 for 20 epochs. The learning rate is selected via hyperparameter tuning from {1e-4, 5e-5, 1e-5}, with 5e-5 yielding the best performance. 

\end{document}